\documentclass[letterpaper, 10 pt, conference]{ieeeconf}  

\usepackage{censor}
\IEEEoverridecommandlockouts                              

\usepackage{amsmath}
\usepackage{amssymb}
\usepackage{url}
\usepackage{comment}
\usepackage{adjustbox}
\usepackage{hyperref}
\hypersetup{
    colorlinks=true,
    linkcolor=blue,
    urlcolor=blue,
    breaklinks=true
}

\usepackage{graphicx} 

\title{\LARGE \bf
An Open-Source Robotics Research Platform for Autonomous Laparoscopic Surgery
}

\author{Ariel Rodriguez$^{\dagger,1}$, Lorenzo Mazza$^{\dagger,1}$, Martin Lelis$^{\dagger,1}$,\\ 
Rayan Younis$^{2,3}$, Sebastian Bodenstedt$^{1}$, 
Martin Wagner$^{2,3}$, Stefanie Speidel$^{1,2}$
\thanks{Affiliations redacted for review}
\thanks{$\dagger$ These authors contributed equally to this work.}
\thanks{$^{1}$
Department of Translational Surgical Oncology, NCT/UCC Dresden, a partnership between DKFZ, MF UKD, TUD, HZDR Dresden, Germany}%
\thanks{$^{2}$ Cluster of Excellence-CeTI, TUD, Germany}%
\thanks{$^{3}$ Department of Visceral, Thoracic and Vascular Surgery, Faculty of Medicine and University Hospital Carl Gustav Carus, TUD, Germany}%
}

\begin{document}

\maketitle
\thispagestyle{empty}
\pagestyle{empty}

\begin{abstract}
Autonomous robot-assisted surgery demands reliable, high-precision platforms that strictly adhere to the safety and kinematic constraints of minimally invasive procedures. Existing research platforms, primarily based on the da Vinci Research Kit, suffer from cable-driven mechanical limitations that degrade state-space consistency and hinder the downstream training of reliable autonomous policies. We present an open-source, robot-agnostic Remote Center of Motion (RCM) controller based on a closed-form analytical velocity solver that enforces the trocar constraint deterministically without iterative optimization. The controller operates in Cartesian space, enabling any industrial manipulator to function as a surgical robot. We provide implementations for the UR5e and Franka Emika Panda manipulators, and integrate stereoscopic 3D perception. We integrate the robot control into a full-stack ROS-based surgical robotics platform supporting teleoperation, demonstration recording, and deployment of learned policies via a decoupled server–client architecture.\footnote{Project page: \url{https://surgical-robot-testbed.github.io/iros_testbed_paper/}} We validate the system on a bowel grasping and retraction task across phantom, ex vivo, and in vivo porcine laparoscopic procedures. RCM deviations remain sub-millimeter across all conditions, and trajectory smoothness metrics (SPARC, LDLJ) are comparable to expert demonstrations from the JIGSAWS benchmark recorded on the da Vinci system. These results demonstrate that the platform provides the precision and robustness required for teleoperation, data collection and autonomous policy deployment in realistic surgical scenarios.
\end{abstract}
\begin{figure}[t]
    \centering
\adjincludegraphics[width=0.7\columnwidth,angle=-90,trim={0 {0.07\height} {0.3\width} 0},clip,cfbox=white 0pt 0pt]{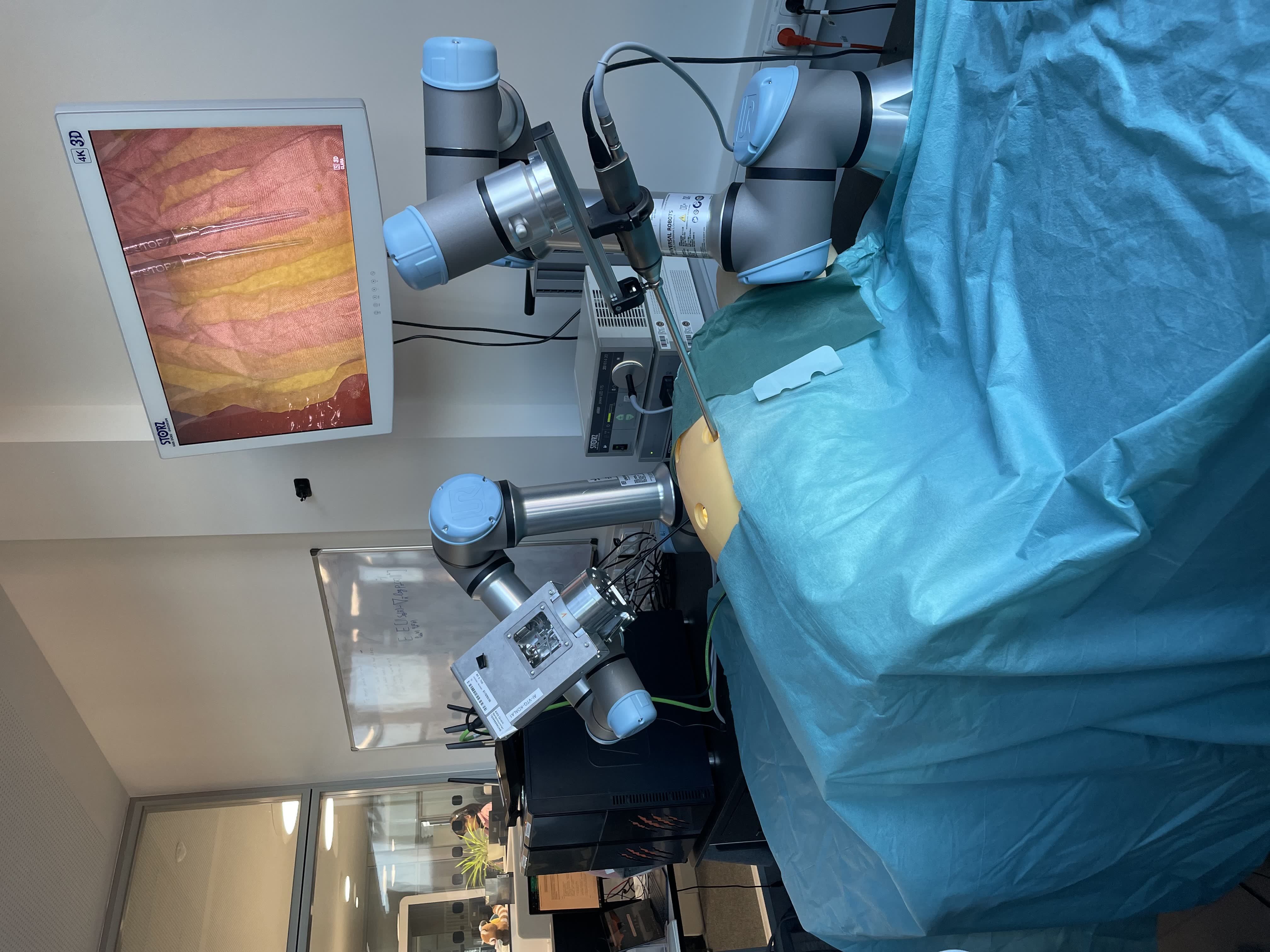}\\[-1pt]
\adjincludegraphics[width=0.7\columnwidth,trim=0 0 0 {0.5\height},clip]{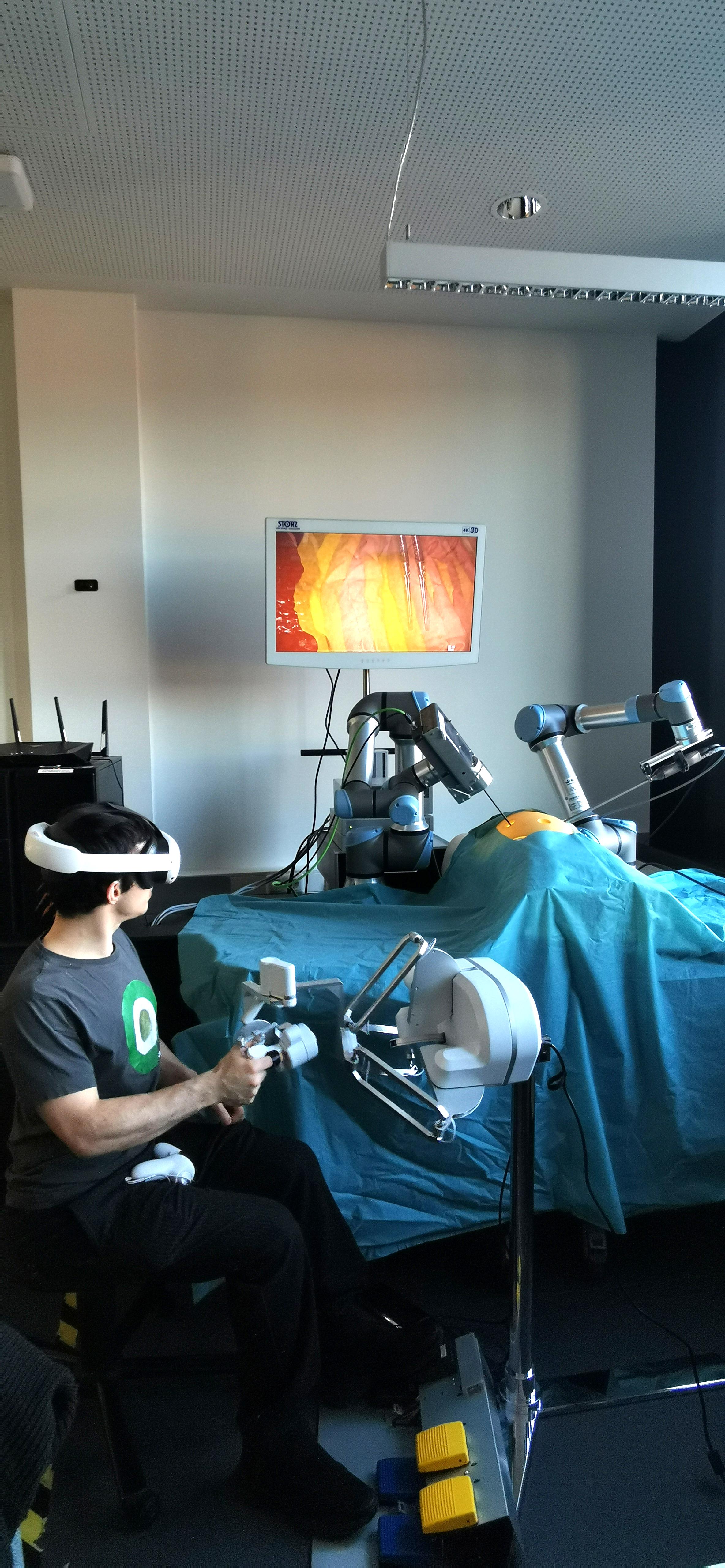}
\caption{System overview of the platform: the teleoperation interface consists of a haptic input device and a clutch pedal board, combined with a Meta Quest headset providing immersive 3D perception. On the patient side, the hardware setup features two UR5e robotic arms, one for controlling the surgical instrument and the other equipped with a stereoscopic endoscope to capture the surgical workspace.}
\label{fig_cover}
\end{figure}

\section{INTRODUCTION}

Robot-Assisted Surgery (RAS) has revolutionized procedural standards, offering patients shorter hospital stays, faster recovery times, and fewer reoperations, while providing surgeons with improved dexterity and ergonomics~\cite{dupont2021decade}. Beyond immediate clinical benefits, these systems offer a critical pathway to address the projected global surgeon shortage~\cite{perera2021global} by enabling research into the automation of surgical subtasks. Imitation learning approaches~\cite{kim2024srt, kim2025srth, mazza2026moe} have demonstrated the feasibility of automating surgical tasks such as tissue manipulation, multi-step dissection procedures, and bowel grasping and retraction. However, the successful translation of these algorithms for future clinical use requires platforms that are extremely precise and strictly adhere to the constraints of laparoscopy. 
\bigskip

Currently, the lack of an accessible and standardized hardware platform presents a major barrier to surgical automation research. The da Vinci Surgical System (Intuitive Surgical, Inc., Sunnyvale, CA, USA) is the most widely deployed robotic platform for RAS in clinical practice, and it accounts for the majority of robotic procedures worldwide \cite{intuitive2024esg}.
However, modern clinical systems are protected by strict proprietary software architectures that prevent the low-level control access required for training autonomous policies. 
Consequently, the academic community has relied primarily on the first-generation da Vinci Research Kit (dVRK)~\cite{dettorre2021accelerating}, assembled from retired clinical hardware, and the Raven-II~\cite{hannaford2013raven}, which provided one of the first open-architecture cable-driven systems for research. These platforms have enabled significant advances in perception and subtask automation; however, they rely on cable-driven transmission mechanisms designed primarily for human-in-the-loop teleoperation. As noted in recent studies on automation~\cite{cui2023caveats}, the inherent mechanical compliance, backlash, and hysteresis of these aging cable-driven systems introduce significant kinematic uncertainty.
This mechanical noise degrades state-space consistency, resulting in the robot’s failure to perform even simple visual-servoing tasks~\cite{hwang2020efficiently}.   
In response, recent works like the Smart Tissue Autonomous Robot (STAR)~\cite{shademan2016supervised} and MOPS~\cite{schwaner2021mops} have demonstrated that industrial collaborative robots (e.g., KUKA, Universal Robots) offer the stiffness and precision necessary for high-fidelity automation, provided they are controlled effectively to respect surgical constraints.

\medskip
A fundamental requirement for any such laparoscopic system is the Remote Center of Motion (RCM) constraint, which ensures that instruments pivot about a fixed point at the trocar insertion site to prevent tissue trauma~\cite{zhang2024rcm}. RCM implementation is generally categorized into mechanical and software-based approaches~\cite{sandoval2018comparison}. 

 Mechanical RCMs, such as the remote kinematic center of the da Vinci system, physically constrain motion via a parallelogram linkage. While safety is intrinsic to the mechanism, these designs are bulky, mechanically complex, and restrict the workspace.

Software-based RCMs offer greater flexibility by enforcing the constraint algorithmically, allowing general-purpose industrial arms to function as surgical assistants. Early approaches utilized redundant degrees of freedom to decouple the pivoting motion from the surgical task using null-space projection~\cite{azimian2010constrained}. More recently, optimization-based methods, such as Quadratic Programming (QP)~\cite{kapoor2006constrained}, have been employed to handle RCM constraints alongside joint limits and collision avoidance.  While effective, these methods often require heuristic parameter tuning and can introduce potential discontinuities near singularities.  
\medskip

To address these hardware and control limitations, we present a surgical robotics research platform built around a UR5e (Universal Robots, Odense, Denmark) manipulator and a novel open-source robot agnostic RCM controller. Unlike optimization-based solvers, our system utilizes a closed-form analytical velocity controller. By analytically deriving the required end-effector twist via vector algebra, our approach enforces the RCM constraint deterministically, ensuring smooth velocity profiles essential for high-quality imitation learning data. The system is based on a full-stack ROS-based software architecture for teleoperation, demonstration recording, and policy deployment. Additionally, the platform incorporates a stereoscopic endoscope (TIPCAM1 S 3D, Karl Storz SE \& Co. KG, Tuttlingen, Germany) to provide camera feedback and optionally supports the integration of real-time immersive 3D perception~\cite{mats2025endomersion} via a Meta Quest headset (Meta Platforms Inc., Menlo Park, CA, USA).
\medskip

Validating the fidelity of such a platform requires quantitative metrics that extend beyond task completion. Motion smoothness, a performance metric used in
human motor control research that correlates with healthy and
coordinated movement~\cite{beck2018sparc}, is a well-established proxy for trajectory quality
in surgical skill assessment. Smoother instrument motions reduce the risk of unintended tissue damage caused by jerky movements or tremor~\cite{aghazadeh2023motion}, and smoothness-based metrics have been shown to correlate significantly with surgical experience level across laparoscopic, endovascular, and robotic platforms~\cite{murali2021velocity, jantscher2018toward}. Therefore, we evaluate our platform's trajectory quality against clinical benchmarks using established smoothness metrics across phantom, \textit{ex vivo}, and \textit{in vivo} porcine laparoscopic procedures, demonstrating that the platform works robustly in realistic surgical scenarios and provides the necessary fidelity for next-generation automation research.

\vspace{5mm}

Our major contributions are the following:
\begin{itemize}
\item An open-source robot-agnostic RCM controller based on a closed-form analytical solver that enforces kinematic constraints via Cartesian velocities, ensuring deterministic execution without the need for iterative optimization.
\item A full-stack ROS-based software architecture for teleoperation, data recording, and deployment of deep learning autonomous policies.
\item Comprehensive experimental validation across phantom, \textit{ex vivo}, and \textit{in vivo} porcine procedures, demonstrating that the platform maintains sub-millimeter RCM deviations and expert-level smoothness under realistic surgical conditions, confirming its readiness for translational surgical autonomy research.
\end{itemize}
Code will be made available upon acceptance\textsuperscript{1}.

\section{SYSTEM OVERVIEW}

\begin{figure*}[ht]
    \includegraphics[width=\textwidth]{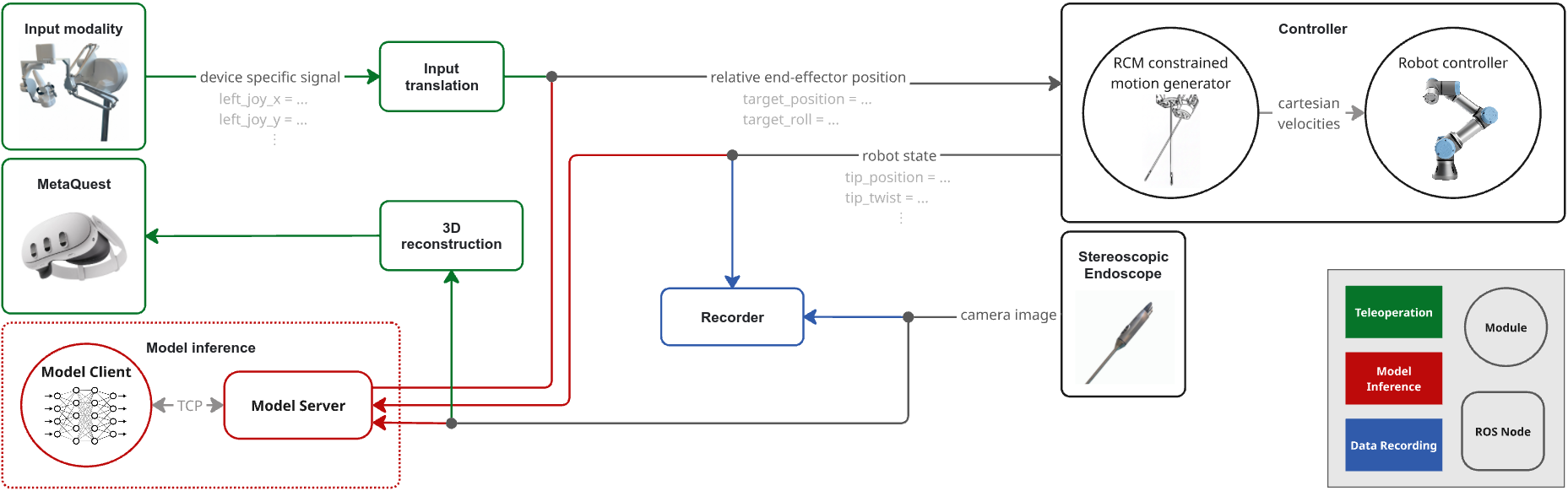}
    \caption{The platform is built with modularity and scalability at its core, with inter-component communication abstracted through ROS topics to remain hardware-agnostic. This architecture allows input devices and robot actuators to be readily exchanged, and new modalities can be incorporated by simply implementing the corresponding nodes or C++ modules.\\
    In teleoperation mode, the operator's input device supplies target position commands to the robot controller, while stereoscopic 3D visualization is rendered for the operator through the Meta Quest headset. During policy deployment, the control signal is instead generated by the autonomous policy, which interfaces with the ROS ecosystem through a ZMQ-based server–client architecture, keeping model inference decoupled from the broader system. \cite{meta_quest3, ur_ur5e}
    }
    \label{fig_overview}
\end{figure*}

\subsection{ROS framework}
We employ the ROS (Robot Operating System) framework to modularize the platform's control software. By relying on ROS nodes we decouple the platform's modules from one another. This allows to exchange input devices or robot actuators without modifying the rest of the system. An overview of the entire software stack is shown in Fig. \ref{fig_overview}.
The RCM-constrained robot controller operates in Cartesian space and is therefore agnostic to the specific hardware. Any robot arm capable of executing the Cartesian velocity commands issued by the controller can be used. We currently provide implementations for the Franka Emika Panda and the Universal Robots UR5e arms. 
We implement ROS nodes that allow teleoperation of the robot arm using either a Lambda.7 haptic device (Force Dimension, Nyon, Switzerland) or a joystick-based controller. Input devices and endoscope nodes may run on separate machines that connect to a ROS master server, allowing operators to control the robot from a different and potentially remote workstation. To ensure temporal consistency across all machines involved, we synchronize clocks using the Precision Time Protocol (PTP), with the machine running the robot controller serving as the master clock.
\subsection{Data Collection}
During teleoperation, we record raw sensor data into CSV files, capturing observations from all desired ROS topics. When a topic streams frames from an endoscope, the recording node saves them as compressed MP4 videos, significantly reducing the required storage size compared to raw image sequences. These recordings are later post-processed offline to produce structured datasets suitable for policy learning. This separation of raw data collection from dataset generation allows adapting the post-processing pipeline to different learning frameworks and data formats without modifying the recording infrastructure.
\subsection{Policy Deployment}
For policy roll-outs, we decouple the inference logic from ROS using a ZMQ-based \cite{zeromq} server–client architecture. An intermediate model server node serves as the interface layer between the learned policy and the robot controller. It aggregates and synchronizes the necessary observation data from the system, forwards it to the policy client, and relays the predicted control signals back to the robot controller. Additionally, the model server incorporates a safety controller that validates the actions predicted by the autonomous policy, ensuring that only safe commands are forwarded to the robot. This architecture is environment-agnostic: any imitation learning or reinforcement learning policy can serve as the client, fully independent from the server-side environment. As a result, policies can be rolled out across on-hardware (\textit{in vivo}, \textit{ex vivo}, phantom) and simulation environments without modifying the infrastructure, maintaining a consistent setup throughout.

\subsection{Remote Center of Motion}
We implement a velocity-level controller that maintains a Remote Center of Motion (RCM) constraint while moving the instrument as desired. The RCM point corresponds to the trocar insertion point, which must remain stationary throughout the procedure. We present two complementary control strategies that differ in their input parameterization while maintaining the same RCM constraint.

Consider a robotic manipulator with end-effector frame positioned at $\mathbf{p}_{ee} \in \mathbb{R}^3$ in the robot base frame. A laparoscopic instrument is rigidly attached to the end-effector flange, with its shaft passing through a fixed RCM point $\mathbf{p}_{rcm} \in \mathbb{R}^3$ and terminating at the instrument tip positioned at $\mathbf{p}_{tip} \in \mathbb{R}^3$. All positions are expressed in the robot base frame.

We define a vector from the end-effector to the RCM point as:
\begin{equation}
\mathbf{r}_{ee} = \mathbf{p}_{rcm} - \mathbf{p}_{ee}
\end{equation}

We additionally define an instrument shaft vector from the RCM point to the tip:
\begin{equation}
\mathbf{r}_{shaft} = \mathbf{p}_{tip} - \mathbf{p}_{rcm}
\end{equation}

\subsubsection{Cartesian Tip Velocity Control}
In this approach, the input corresponds to the desired instrument tip velocities $\mathbf{v}_{tip} \in \mathbb{R}^3$ (translational) and $\omega_{roll} \in \mathbb{R}$ (roll about the shaft axis). The controller computes the required end-effector velocities $\mathbf{v}_{ee}$ and $\boldsymbol{\omega}_{ee}$ that satisfy the RCM constraint.

The RCM constraint requires that the instrument shaft always passes through $\mathbf{p}_{rcm}$, implying that any motion of the tip tangent to the shaft must induce a pure rotation of the shaft about the RCM point. The required angular velocity component to achieve this is:
\begin{equation}
\boldsymbol{\omega}_{pivot} = \frac{\mathbf{r}_{shaft} \times \mathbf{v}_{tip}}{\lVert\mathbf{r}_{shaft}\rVert^2}
\end{equation}

The desired roll motion about the shaft axis contributes an angular velocity component along the shaft direction:
\begin{equation}
\boldsymbol{\omega}_{roll} = \omega_{roll} \cdot \hat{\mathbf{r}}_{shaft}
\end{equation}
where $\hat{\mathbf{r}}_{shaft} = \mathbf{r}_{shaft} / \lVert\mathbf{r}_{shaft}\rVert$ is the unit vector along the shaft.

The total required angular velocity of the end-effector is:
\begin{equation}
\boldsymbol{\omega}_{ee} = \boldsymbol{\omega}_{pivot} + \boldsymbol{\omega}_{roll}
\end{equation}

For the linear velocity, the desired tip motion can be decomposed into components parallel and perpendicular to the shaft. The parallel component corresponds to insertion/retraction:
\begin{equation}
\mathbf{v}_{insertion} = \left(\mathbf{v}_{tip} \cdot \hat{\mathbf{r}}_{shaft}\right) \hat{\mathbf{r}}_{shaft}
\end{equation}

The required end-effector linear velocity must account for both the insertion motion and the effect of the angular velocity on the point offset by $\mathbf{r}_{ee}$:
\begin{equation}
\mathbf{v}_{ee} = \mathbf{v}_{insertion} - \boldsymbol{\omega}_{ee} \times \mathbf{r}_{ee}
\end{equation}

This control mode allows the operator to move the tip of the instrument inside the patient body using Cartesian coordinates, and to rotate it around its own axis.

\begin{figure}[t]
    \centering
    \adjincludegraphics[width=0.7\columnwidth,trim=0 0 0 {0.1\height},clip]{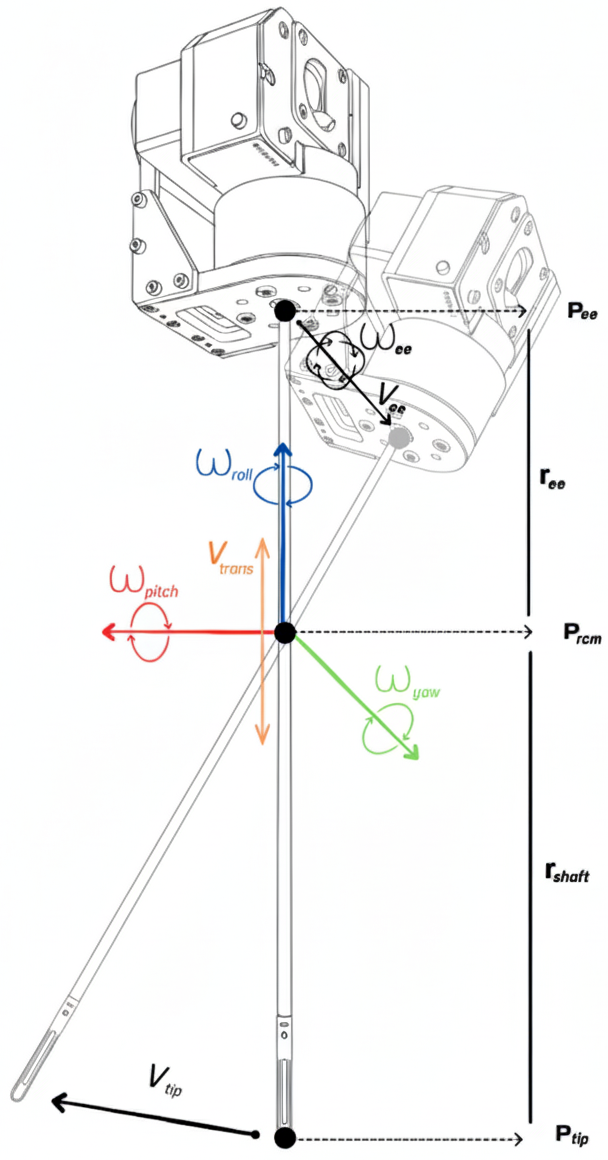}
\caption{Remote-Center-of-Motion instrument control showing spherical coordinates (pitch, yaw, roll, translation) and Cartesian tip velocity methods, with all vectors expressed in the instrument robot's base frame.}
\label{fig_rcm}
\end{figure}

\subsubsection{Spherical Coordinate Control}

An alternative parameterization uses spherical coordinates that directly correspond to the surgical degrees of freedom. In this approach, the input parameters correspond to:
\begin{itemize}
\item $\omega_{pitch} \in \mathbb{R}$: angular velocity for elevation angle (up/down pivoting)
\item $\omega_{yaw} \in \mathbb{R}$: angular velocity for azimuthal angle (left/right pivoting)
\item $\omega_{roll} \in \mathbb{R}$: angular velocity around the instrument shaft axis
\item $v_{trans} \in \mathbb{R}$: linear velocity for insertion/retraction along the shaft
\end{itemize}

All angular velocities are expressed in the base frame. The end-effector angular velocity is constructed as:
\begin{equation}
\boldsymbol{\omega}_{ee} = \begin{bmatrix} \omega_{pitch} \\ \omega_{yaw} \\ 0 \end{bmatrix} + \omega_{roll} \cdot \hat{\mathbf{r}}_{ee}
\end{equation}
where $\hat{\mathbf{r}}_{ee} = \mathbf{r}_{ee} / \lVert\mathbf{r}_{ee}\rVert$ is the unit vector along the instrument shaft direction from the end-effector to the RCM point.

The pitch and yaw components directly control the pivoting motion around the RCM point, while the roll component rotates the instrument about its own axis. The end-effector linear velocity is then:
\begin{equation}
\mathbf{v}_{ee} = v_{trans} \cdot \hat{\mathbf{r}}_{ee} - \boldsymbol{\omega}_{ee} \times \mathbf{r}_{ee}
\end{equation}

This formulation allows controlling the instrument using the roll, pitch, yaw angles and depth translation.

\subsection{3D Perception}
Our system employs a TIPCAM1 S 3D stereo endoscope (Karl Storz SE \& Co. KG, Tuttlingen, Germany) mounted on a second robotic arm to provide stereoscopic visualization. The stereo video feed is displayed on either a 3D monitor or a Meta Quest headset \cite{mats2025endomersion}, enabling depth perception during task execution and demonstration recording. 

The camera frame is determined using the forward kinematics of the camera-holding arm combined with a measured offset from its end-effector to the endoscope tip. Since the transformation between the two robot bases is known, we can transform instrument commands from the camera frame to the instrument robot frame. This control ensures surgeon commands remain intuitive regardless of viewing angle, providing natural teleoperation.

\begin{figure}[t]
    \centering
    \includegraphics[width=\columnwidth]{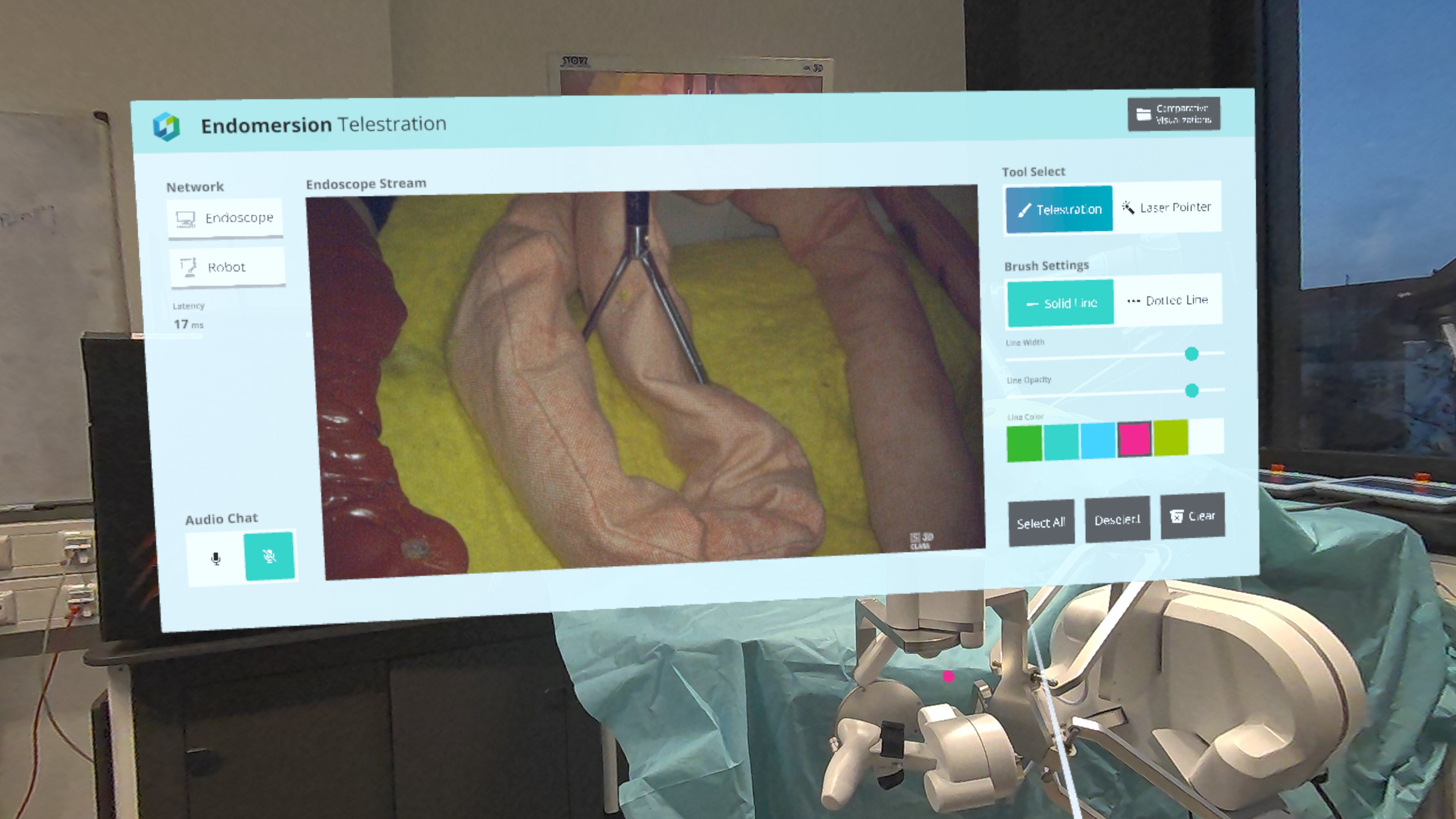}
    \caption{Immersive stereoscopic visualization through the Meta Quest headset running Endomersion~\cite{mats2025endomersion} showing the surgical workspace captured by the endoscope. Only the left eye view is shown; the right eye receives the corresponding stereo pair to enable depth perception.}
    \label{fig_endomersion}
\end{figure}
  \begin{figure*}[t]
      \centering
      \includegraphics[]{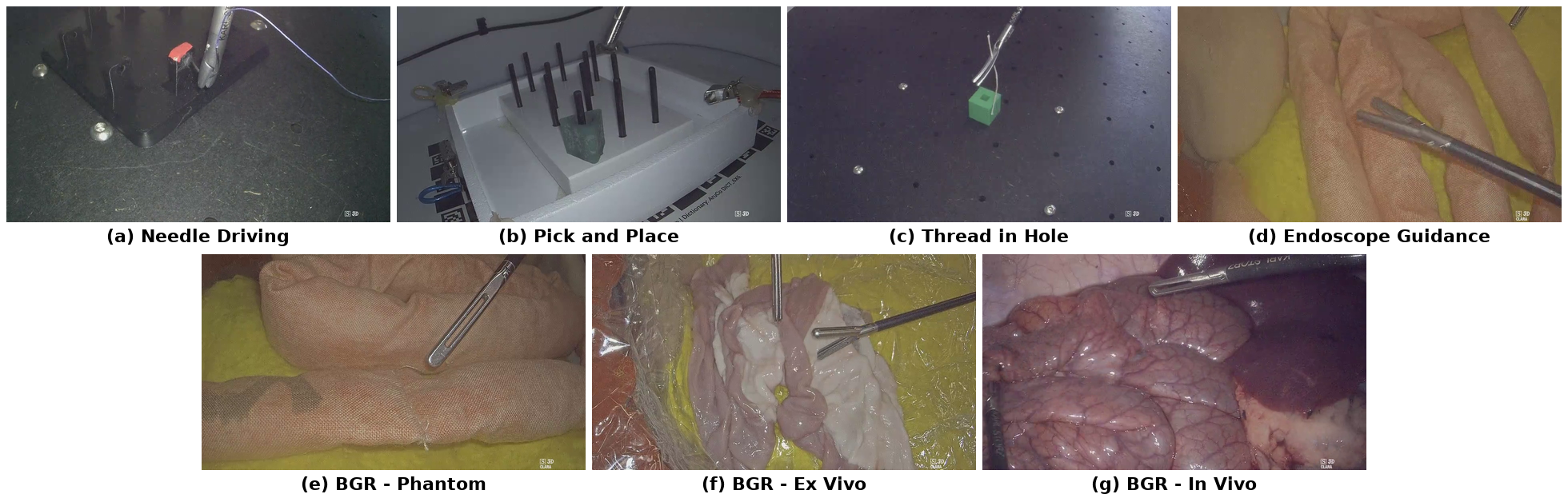}
      \caption{Representative frames from datasets collected using the platform: (a)~Needle Driving, (b)~Pick and Place, (c)~Thread in Hole, (d)~Endoscope Guidance, and Bowel
  Grasping and Retraction~(BGR) across (e)~phantom, (f)~\textit{ex vivo}, and (g)~\textit{in vivo} settings.}
      \label{fig_datasets}
  \end{figure*}

\subsection{Velocity Profiler}
The controller accepts desired tip positions or desired angular orientations and insertion depth as input. To generate smooth motion toward the target, we employ fifth-order polynomial velocity profilers for each degree of freedom. At each control cycle, when a new desired position is commanded, the profiler generates a trajectory that smoothly transitions from the current state (position, velocity, and acceleration) to the target position over a duration of \(d\) seconds. The quintic polynomial formulation ensures continuous position, velocity, and acceleration profiles while satisfying boundary conditions at both endpoints \cite{lynch2017modern, craig2009introduction}. The profiler enforces user-defined jerk and acceleration limits by automatically extending \(d\) if kinematic constraints would otherwise be violated. We set \(d\) to a base value of 0.5 seconds in our system. At each timestep, the profiler outputs the instantaneous velocity command $v_i(t)$, which is then passed to the RCM controller to compute the corresponding end-effector velocities.

\section{APPLICATION AND PLATFORM EVALUATION}
\subsection{Data Collection}
The platform's modular architecture supports data collection across a variety of surgical tasks. Figure~\ref{fig_datasets} shows representative frames from the datasets recorded using our system. \textit{Needle Driving}, \textit{Pick and Place}, and \textit{Thread in Hole} are surgical training exercises performed on 3D-printed boards:  \textit{Needle Driving}  involves driving a surgical needle into a 2\,mm hole,  \textit{Pick and Place} requires grasping a peg from one pole and transferring it to another, and  \textit{Thread in Hole} involves guiding a surgical thread into the hole of a 3D-printed figure. \textit{Endoscope Guidance} uses the RCM controller on the camera-holding arm to autonomously navigate the endoscope and maintain optimal visualization of the surgical field. 

\medskip
For quantitative evaluation, we focus on the Bowel Grasping and Retraction~(BGR) task, a four-phase collaborative procedure in which the controlled instrument operates
alongside a human-controlled assistant instrument. In the first phase, the assistant identifies a target grasping location on the bowel, which the controlled instrument must navigate to
and grasp. In the second phase, the grasp is held while the assistant acquires a separate hold on the bowel. The third phase requires retracting the bowel so that the tissue is
tensioned between both instruments and clearly visible in the camera view. Finally, in the fourth phase, both the grasp and the instrument positions must be held steadily. This
task is evaluated in three settings: a surgical phantom~\cite{openhelp}, an \textit{ex vivo} environment, and an \textit{in vivo} environment~\cite{openh2025datacollection}.

To validate both the quality of data collected within our platform and its suitability for policy deployment, we train a policy via imitation learning on the collected dataset. The model achieves a success rate of 85\% (17/20 roll-outs), underscoring the platform's robustness as a foundation for policy learning\cite{mazza2026moe}.

\subsection{Evaluation Metrics}
We quantitatively evaluate the platform across the collected Bowel Grasping and Retraction dataset, focusing on the two fundamental metrics for a laparoscopic research platform: \textit{RCM Constraint Adherence} and \textit{Trajectory Smoothness}.

\paragraph{RCM Constraint Adherence}
To quantify the precision of our kinematic constraints, we measure the physical RCM error at each control timestep ($\sim$500\,Hz) based on the recorded UR5e joint states. At initialization, we calibrate the instrument shaft vector relative to the end-effector frame to account for the lateral offset of the custom tool holder.  At each subsequent timestep, the shaft axis is reconstructed in world coordinates by applying the current flange orientation to the calibrated direction. The RCM deviation is computed as the minimum Euclidean distance between the commanded pivot point and this reconstructed infinite shaft line. We report the mean and maximum deviation per episode in millimeters.

\begin{figure}[t]
    \includegraphics[width=\columnwidth]{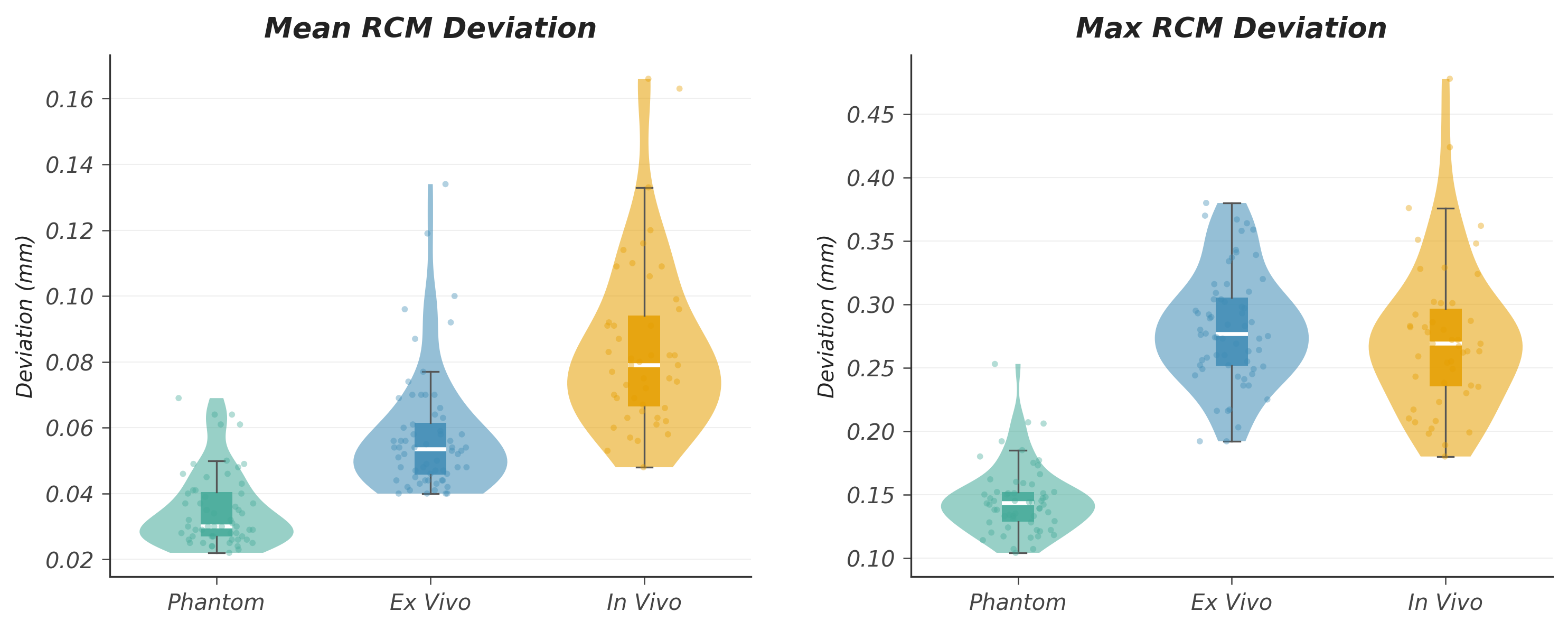}%
    \caption{Remote Center of Motion (RCM) deviation across experimental conditions. Mean deviation (left) and worst-case maximum deviation (right) remain sub-millimeter across all conditions, with phantom ($\tilde{x}=0.035$\,mm), ex vivo ($\tilde{x}=0.052$\,mm), and in vivo ($\tilde{x}=0.079$\,mm) showing a progressive increase consistent with greater tissue interaction complexity.}
    \label{fig_rcm_deviation}
\end{figure}

Figure \ref{fig_rcm_deviation} demonstrates successful RCM constraint maintenance across all three experimental scenarios. Both mean and maximum deviations remain sub-millimetric in phantom, ex-vivo, and in-vivo experiments. In-vivo trials exhibit slightly larger deviations due to tissue compliance and pressure forces at the trocar insertion point, which cause small displacements of the nominal RCM location. Overall, these results demonstrate effective RCM constraint enforcement.

\begin{figure}[t]
    \centering
    \includegraphics[width=\columnwidth]{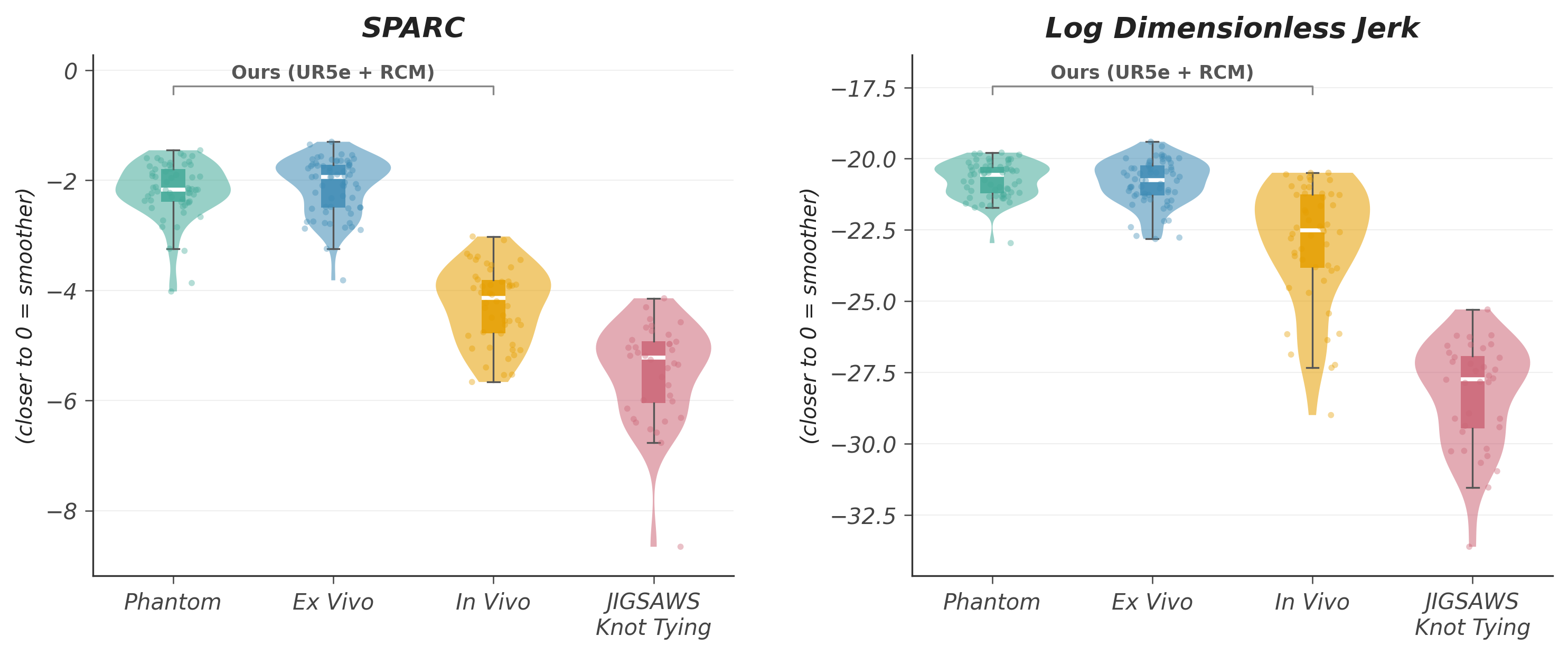}%
    \caption{Trajectory smoothness comparison using SPARC (left) and Log Dimensionless Jerk (right) metrics, where values closer to zero indicate smoother motion. Our robot testbed produces progressively less smooth trajectories from phantom to in vivo, consistent with increasing difficulty for the Bowel Grasping and Retraction task. All three conditions yield comparable trajectory smoothness metrics to the JIGSAWS Knot Tying dataset recorded on the da Vinci surgical system. All signals downsampled to 5\,Hz for fair comparison.}
    \label{fig_smoothness}
\end{figure}

\paragraph{Trajectory Smoothness}
To assess the kinematic quality of the platform, we analyze the 3D instrument tip velocity profiles using two complementary metrics established in motor control literature: Spectral Arc Length (SPARC) and Log Dimensionless Jerk (LDLJ).
SPARC~\cite{balasubramanian2011robust} quantifies smoothness by calculating the arc length of the normalized Fourier magnitude spectrum of the velocity profile. Smooth, bell-shaped speed profiles produce compact spectra with arc lengths closer to zero, while jerky, intermittent movements introduce high-frequency components that result in more negative values. SPARC is dimensionless and robust to measurement noise.

LDLJ~\cite{balasubramanian2015analysis} integrates the squared jerk (third time derivative of position) over the movement duration, normalized by peak speed to ensure dimensionlessness. The logarithmic scale improves sensitivity to intermittency. More negative LDLJ values indicate jerkier motion. Unlike
unnormalized jerk measures, which conflate smoothness with movement
duration and path length, LDLJ isolates the intermittency component
and can reliably distinguish surgical skill
levels~\cite{aghazadeh2023motion}.

We compute these metrics on the tip velocity profiles downsampled to 5\,Hz. To contextualize our results, we apply the same evaluation pipeline to the JIGSAWS Knot Tying dataset~\cite{ahmidi2017dataset} (using PSM1 kinematics). JIGSAWS
is one of the most widely used open-source benchmarks for expert
surgical kinematics. Our aim is not to directly compare the performance of the two platforms,
since the tasks differ in nature. Rather, we use JIGSAWS as a
reference to verify that the smoothness metrics produced by our
platform fall within the range expected of an established surgical
robotics system.

Figure \ref{fig_smoothness} presents the smoothness measurements across all experimental scenarios. Phantom and ex-vivo experiments exhibit nearly identical and excellent smoothness characteristics, with SPARC values around -2 and LDLJ values around -20. In-vivo experiments exhibit moderately reduced smoothness (SPARC $\approx -4$, LDLJ $\approx -22$) due to forces exerted by the trocar on the abdominal wall. Notably, all three experimental conditions demonstrate comparable smoothness to the JIGSAWS expert demonstrations (SPARC $\approx -5$, LDLJ $\approx -27$).

\section{CONCLUSION}

We have presented a robot-agnostic, open-source robotics platform for autonomous laparoscopic surgery that addresses drawbacks of current systems through an analytical RCM controller deployed in a ROS architecture. Quantitative validation across phantom, ex vivo, and in vivo scenarios demonstrates sub-millimeter RCM deviations and trajectory smoothness metrics comparable to expert performance on the da Vinci system, validating that industrial robots can provide the precision required for surgical automation research.

A key advantage of industrial robots is their modular end-effector interface, which enables straightforward integration of commercial force-torque sensors between the flange and surgical tool. Future work will incorporate a 6-axis force sensor to capture the full wrench at the robot flange. This will enable haptic feedback for teleoperation and proprioceptive data for learning force-aware autonomous policies.
Additionally, the platform will enable diverse data collection and evaluation of surgical foundation models requiring cross-scenario generalization.

\bibliographystyle{IEEEtran}
\bibliography{references}

\end{document}